\documentclass[runningheads]{llncs}

\usepackage{graphicx}

\usepackage{tikz}
\usepackage{comment}
\usepackage{amsmath,amssymb} %
\usepackage{color}

\usepackage{booktabs}
\usepackage{epsfig}
\usepackage{graphicx}
\usepackage{float}
\usepackage[export]{adjustbox}
\usepackage{verbatim}
\usepackage{xr}
\usepackage[utf8]{inputenc}
\usepackage{multirow}
\usepackage{caption}
\usepackage{subcaption}
\usepackage{makecell}
\usepackage{hyperref}

\usepackage[normalem]{ulem}

\def\arraystretch{1.2}
\graphicspath{{./figures}}

\begin{document}
\pagestyle{headings}
\mainmatter
\def\ECCVSubNumber{3498}  %

\title{Model-based occlusion disentanglement\\ for image-to-image translation} %

\titlerunning{Model-based occlusion disentanglement for image-to-image translation}
\author{Fabio Pizzati\inst{1,2} \and
Pietro Cerri\inst{2}\and
Raoul de Charette\inst{1}\thanks{Corresponding author.}}
\authorrunning{F. Pizzati et al.}
\institute{Inria, Paris, France\\
\email{\{fabio.pizzati, raoul.de-charette\}@inria.fr} \and
VisLab, Parma, Italy\\
\email{pcerri@ambarella.com}\\
}

\index{de Charette, Raoul}

\maketitle

\begin{abstract}
Image-to-image translation is affected by entanglement phenomena, which may occur in case of target data encompassing occlusions such as raindrops, dirt, etc. Our unsupervised model-based learning disentangles scene and occlusions, while benefiting from an adversarial pipeline to regress physical parameters of the occlusion model.
The experiments demonstrate our method is able to handle varying types of occlusions and generate highly realistic translations, qualitatively and quantitatively outperforming the state-of-the-art on multiple datasets.
\end{abstract}

\begin{figure}
	\centering
	\setlength{\tabcolsep}{0.001\linewidth}
	\renewcommand{\arraystretch}{0.3}
	\tiny
	\begin{tabular}{ccccc}
	     \includegraphics[width=0.198\linewidth]{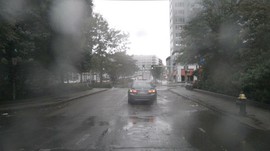}
	     & \includegraphics[width=0.198\linewidth]{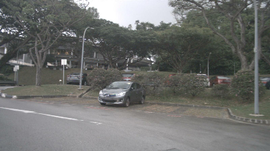}
	     & %
	     \includegraphics[width=0.1985\linewidth]{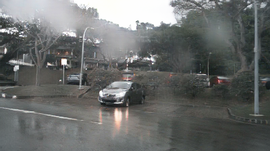}
	     & \includegraphics[width=0.198\linewidth]{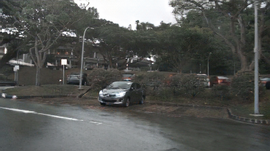}
	     & \includegraphics[width=0.198\linewidth]{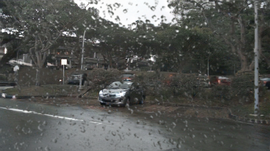}\\
	     Target sample & Source & MUNIT~\cite{huang2018multimodal} & \textbf{Ours disentangled} & \textbf{Ours w/ focus drops}
	\end{tabular}
	\caption{
	Our method learns to disentangle scene from occlusions using unsupervised adversarial disentanglement with guided injection of a differentiable occlusion model. Here, we separate unfocused drops from rainy scene and show that, opposed to existing baselines, we learn a fully disentangled translation without drops and can re-inject occlusions with unseen parameters (e.g. in-focus drops).
	}\label{fig:opening}
\end{figure}

\section{Introduction}
Image-to-image (i2i) translation GANs are able to learn the $\text{source}\mapsto{}\text{target}$ style mapping of paintings, photographs, etc.~\cite{zhu2017unpaired,liu2017unsupervised,isola2017image}. In particular, synthetic to real~\cite{bi2019deep} or weather translation~\cite{qu2019enhanced,shen2019towards,pizzati2019domain} attracted many works as they are alternatives to the menial labeling task, and allow domain adaptation or finetuning to boost performance on unlabeled domains.
However, GANs notoriously fail to learn the underlying physics~\cite{xie2018tempogan}. This is evident when \textit{target} data encompass occlusions (such as raindrop, dirt, etc.) as the network will learn an entangled representation of the scene with occlusions. For example, with $\text{clear}\mapsto{}\text{rain}$ the GAN translation will tend to have too many drops occlusions, often where the translation is complex as it is an easy way to fool the discriminator.

We propose an unsupervised model-based adversarial disentanglement to separate target and occlusions.
Among other benefits, it enables accurate translation to the target domain and permits proper re-injection of occlusions. More importantly, occlusions with different physical parameters can be re-injected, which is crucial since the appearance of occlusions varies greatly with the camera setup. For example, drops occlusions appear different when imaged in-focus or out-of-focus. There are obvious benefits for occlusion-invariant outdoor vision like mobile robotics or autonomous driving.
A comparison showcasing standard i2i (that partially entangles unrealistic drops) and our framework capabilities is available in Fig.~\ref{fig:opening}.
\noindent{}Our method builds on top of existing GAN architectures enabling unsupervised adversarial disentanglement with the only prior of the occlusion model. Parameters of the occlusion model are regressed on the target data and used when training to re-inject occlusions further driven by our disentanglement guide. We demonstrate our method is the only one able to learn an accurate translation in the context of occlusions, outperforming the literature on all tested metrics, and leading to better transfer learning on semantic segmentation. Our method is able to cope with various occlusion models such as drops, dirt, watermark, or else, among which raindrop is thoroughly studied. 
Our contributions may be summarized as follows:
\begin{itemize}
    \item we propose the first unsupervised model-based disentanglement framework,
    \item our adversarial parameter estimation strategy allows estimating and replicating target occlusions with great precision and physical realism,
    \item our disentanglement guidance helps the learning process without losing generative capabilities in the translation task,
    \item we conducted exhaustive experiments on raindrops occlusions proving we outperform the literature, boost transfer learning, and provide a focus agnostic framework of high interest for autonomous driving applications.
\end{itemize}

\section{Related work}
\paragraph{Image-to-image translation.}
Seminal works on image-to-image translation (i2i) was conducted by Isola et al.~\cite{isola2017image} and Zhu et al.~\cite{zhu2017unpaired} for paired and unpaired data respectively, where the later introduced the cycle consistency loss extended in~\cite{zhu2017toward,yi2017dualgan}.
Liu et al.~\cite{liu2017unsupervised} further proposed using Variational Auto Encoders to learn a shared latent space. A common practice to increase accuracy is to learn scene-aware translation exploiting additional supervision from semantic \cite{li2018semantic,ramirez2018exploiting,tang2020multi,cherian2019sem}, instance~\cite{mo2018instagan} or objects~\cite{shen2019towards}.
Furthermore, a recent trend is to use attention-guided translation \cite{mejjati2018unsupervised,ma2018gan,tang2019attention,kim2019u} to preserve important features.\\
Regarding disentangled representations, MUNIT \cite{huang2018multimodal} and DRIT \cite{lee2019drit++} decouple image content and style to enable multi-modal i2i, similar in spirit to our goal, while FUNIT \cite{liu2019few} uses disentangled representations for few-shot learning. FineGAN \cite{singh2019finegan} disentangles background and foreground using bounding boxes supervision.
Multi-domain i2i also opens new directions to control elements at the image level  \cite{choi2018stargan,romero2019smit,anoosheh2018combogan,yang2018crossing,hui2018unsupervised}, since it may be used to represent elements learned from various datasets. Attribute-based image generation \cite{xiao2017dna,xiao2018elegant,zhang2019multi} follows a similar scheme, explicitly controlling features. 
Nonetheless, these methods require attributes annotations or multiple datasets -- hardly compatible with occlusions. Finally, Yang et al. \cite{yang2018towards} exploit a disentangled physical model for dehazing. 

\paragraph{Lens occlusion generation (drops, dirt, soiling, etc.).}
Two strategies co-exist in the literature: physics-based rendering or generative networks. 
Early works on geometrical modeling showcased accurate rendering of raindrops via ray-tracing and 3D surface modeling \cite{roser2009video,roser2010realistic,hao2019learning}, sometimes accounting for complex liquid dynamics \cite{you2015adherent} or focus blur~\cite{roser2010realistic,hao2019learning}. 
A general photometric model was also proposed in \cite{gu2009removing} for thin occluders, while recent works use displacement maps to approximate the raindrops refraction behavior~\cite{porav2019can,alletto2019adherent}.
Generative networks were also recently leveraged to learn general dirt generation~\cite{uricar2019let} but using semantic soiling annotations. To the best of our knowledge, there are no approaches that simultaneously handle occlusions and scene-based modifications with i2i. Note that we intentionally do not review the exhaustive list of works on de-raining or equivalent as it is quite different from disentanglement in the i2i context.

\section{Model-based disentanglement}
\label{sec:method}
We aim to learn the disentangled representation of a target domain and occlusions.
For example, when translating clear to rain images having raindrops on the lens, standard image-to-image (i2i) fails to learn an accurate mapping as the target entangles the scene and the drops on the lens.
We depart from the literature by learning a disentangled representation of the target domain from the injection of an occlusion model, in which physical parameters are estimated from the target dataset. Not only does it allows us to learn the disentangled representation of the scene (e.g. target image without any occlusions) but also to re-inject the occlusion model either with the estimated parameters or with different parameters (e.g. target image with drops in focus).

\begin{figure}[t]
	\centering
	\includegraphics[width=0.9\textwidth]{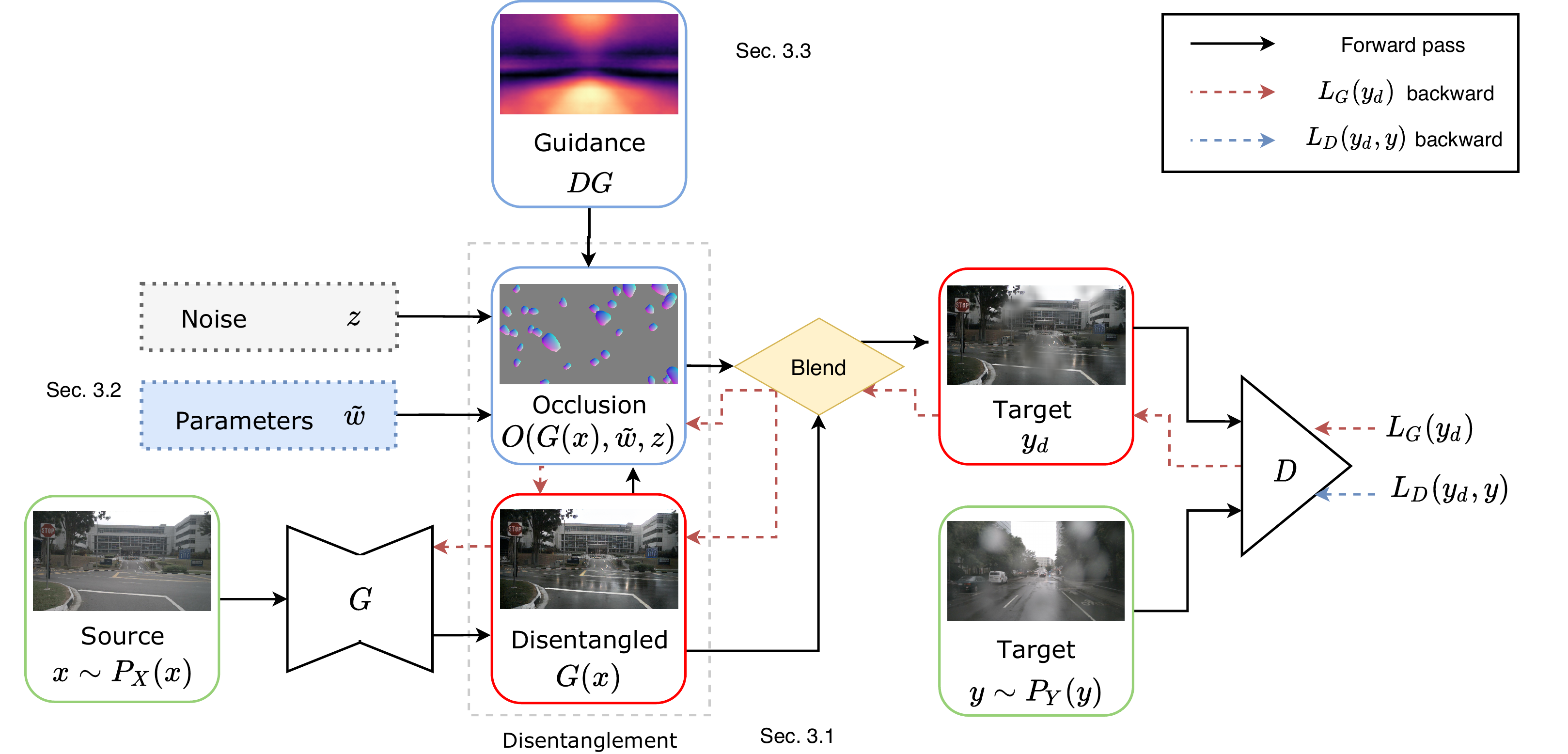}
	\caption{To disentangle the i2i translation process in an unsupervised manner, we inject occlusions $O(.)$ with estimated parameters $\tilde{w}$ before forwarding the generated image $G(x)$ through the discriminator $D$. The Disentanglement Guidance~(DG) avoids losing translation capabilities in low domain shift areas. \textit{Fake} and \textit{real} data are drawn red and green, respectively.}\label{fig:disentanglement}
\end{figure}
Our method handles any sort of occlusions such as raindrops, soil, dirt, watermark, etc. but for clarity we focus on raindrops as it of high interest and exhibits complex visual appearance. 
Fig.~\ref{fig:disentanglement} shows an overview of our training pipeline, which is fully unsupervised and only exploits the prior of occlusion type.
To learn adversarial disentanglement by injecting occlusions~(Sec.~\ref{sec:disentanglement}), we first pretrain a baseline to regress the model parameters~(Sec.~\ref{sec:advmodel}) and estimate domain shifts to further guide the disentanglement learning process~(Sec.~\ref{sec:domainbias}).

\subsection{Adversarial disentanglement} \label{sec:disentanglement}
Let $X$ and $Y$ be the domains of a \textit{source} and a \textit{target} dataset, respectively. 
In an i2i setup, the task is to learn the $X\mapsto{}Y$ mapping translating source to target. 
Now, if the target dataset has occlusions of any sort, $Y$ encompasses two domains: the scene domain $_{S}$, and the occlusion domain $_{O}$.
Formally, as in \cite{pizzati2019domain} we introduce a disentangled representation of domains such that $Y = \{Y_{S}, Y_{O}\}$ and $X=\{X_{S}\}$.
\noindent{}In adversarial training strategies, the generator is led to approximate $P_X$ and $P_Y$, the probability distributions associated with the domains stochastic process, defined as
\begin{equation}
	\begin{split}
	\forall x\in X,x\sim P_X(x),\\
	\forall y\in Y,y\sim P_Y(y).
	\end{split}
\end{equation}

\noindent{}Having occlusions, the target domain $Y$ is interpreted as the composition of two subdomains, and we seek to estimate $P_{Y_S, Y_O}(y_S, y_O)$ corresponding to the scene and occlusion domain. To address this, let's make the \textit{naive} assumption that marginals $P_{Y_S}(y_S)$ and $P_{Y_O}(y_O)$ are independent from each other. Thus, exploiting the definition of joint probability distribution, $P_Y(y)$ becomes 
\begin{equation}
	P_Y(y) = P_{Y_S, Y_O}(y_S, y_O) = P_{Y_S}(y_S)P_{Y_O}(y_O),
	\label{eq:joint}
\end{equation} and it appears that knowing one of the marginals would enable learning disentangled i2i translations between subdomains.
In particular, if $P_{Y_O}(y_O)$ is known it is intuitively possible to learn $X_S\mapsto~Y_S$, satisfying our initial requirement.

In reality, transparent occlusions - such as raindrops - are not fully disentangled from the scene, as their appearance is varying with scene content (see ablation in Sec.~\ref{sec:exp-ablation}). However, the physical properties of occlusions may be seen as quite independent.
As an example, while the appearance of drops on the lens varies greatly with scene background, their physics (e.g. size, shape, etc.) is little- or un- related to the scene. 
Fortunately, there is extensive literature providing appearance models for different types of occlusions (drop, dirt, etc.) given their physical parameters, which we use to estimate $P_{Y_O}(y_O)$.
We thus formalize occlusion models as $O(s, w, z)$ parametrized by the scene $s$, the set of disentangled physical properties $w$, and a random noise vector $z$. The latter is used to map characteristics that can not be regressed as they are stochastic in nature.
This is the case for raindrops positions for example. Assuming we know the type of occlusion such as drop, dirt, etc., we rely on existing models (described in ~Sec.~\ref{sec:exp}) to render the visual appearance of occlusions.

Ultimately, as depicted in Fig.~\ref{fig:disentanglement}, we add occlusions rendered with a known model on generated images before forwarding them to the discriminator. 
Assuming sufficiently accurate occlusion models, the generator $G$ is pushed to estimate the disentangled $P_{Y_S}$, thus learning to translate only scene-related features. As a comparison to a standard LSGAN \cite{mao2017least} training which enforces a zero-sum game by minimizing

\begin{equation}
\begin{split}
y_{\text{d}} &= G(x), \\
L_{\text{gen}} &= L_G(y_d) = \mathbb{E}_{x~\sim P_X(x)}[(D(y_d) - 1) ^ 2], \\
L_{\text{disc}} &= L_D(y_d, y) = \mathbb{E}_{x~\sim P_X(x)}[(D(y_d)) ^ 2] + \mathbb{E}_{y~\sim P_Y(y)}[(D(y) - 1) ^ 2],
\end{split}
\end{equation}

with $L_{\text{gen}}$ and $L_{\text{disc}}$ being respectively the tasks of the generator $G$ and discriminator $D$, we instead learn the desired disentangled mapping by injecting occlusions $O(.)$ on the translated image. Hence, we newly define $y_{\text{d}}$ as the disentangled composition of translated scene and injected occlusions, that is
\begin{equation}
\label{eq:gan_dis_objective}
    \begin{split}
    y_{\text{d}} &= \alpha G(x) + (1 - \alpha)O(G(x), \tilde{w}, z), \\
    \end{split}
\end{equation}
where $\alpha$ is a pixel-wise measure of the occlusion transparency. Opaque occlusions will locally set $\alpha = 1$ while a pixel with transparent occlusions has $\alpha < 1$.
Because physical parameters greatly influence the appearance of occlusions (e.g. drop in focus or out of focus), we render occlusions in Eq.~\ref{eq:gan_dis_objective} using $\tilde{w}$, the optimal set of physical parameters to model occlusions in $Y$.

\begin{figure}
	\centering
	\includegraphics[width=0.75\textwidth]{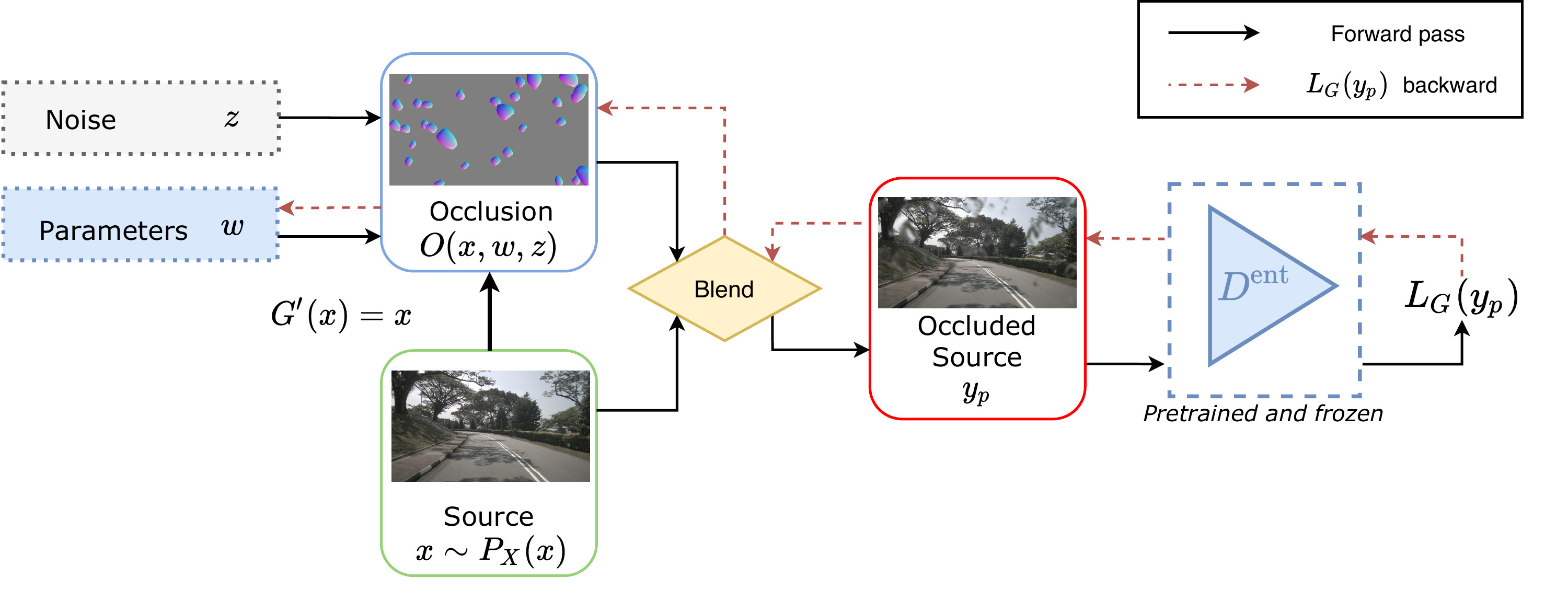}
	\caption{We estimate the optimal parameters to use for the disentanglement adding occlusions on $source$ images and optimizing the parameters of the physical model in order to fool the discriminator. Since we are not using a generator network, the gradient, represented as red arrows, flows only in the occlusion model direction.}\label{fig:advattack}
\end{figure}
\subsection{Adversarial parameters estimation} \label{sec:advmodel}
We estimate the set of physical parameters $\tilde{w}$ from $Y$ in an unsupervised manner, benefiting from the entanglement of scene and occlusion in the target domain. We build here upon Eq.~\ref{eq:joint}. 
Assuming a naive i2i baseline being trained on source and target data, \textit{without} disentanglement, the discriminator learned to distinguished examples from source and target by discriminating $P_X=P_{X_S}$ from $P_Y=P_{Y_S}(y_S)P_{Y_O}(y_O)$.
Let's consider the trivial case where a generator $G'$ performs identity (i.e. $G'(x) = x$) then we get $P_{Y_S}(y_S) = P_{X_S}(x_S)$ and it is now possible to estimate the optimal $\tilde{w}$ as it corresponds to the distance minimization of $P_Y$ and $P_X$. 
Intuitively, as the domain gap results of both occlusion and scene domain (which is fixed), reducing the source and target domain gap implies reducing the occlusion domain gap, in extenso regressing $w$.

Fig.~\ref{fig:advattack} illustrates the estimation process.
In practice, we pretrain a simple i2i baseline (e.g. MUNIT\cite{huang2018multimodal}) to learn $X\mapsto{}Y$ in a naive - entangled - manner, by training a generator and discriminator.
We then freeze the naive discriminator denoted $D^{\text{ent}}$ and solve the following optimization objective
\begin{equation}
\begin{gathered}
\label{eq:param-optim}
    y_p = \alpha G'(x) + (1 - \alpha)O(G'(x), w, z),\\
	\min_{w} L_G(y_p)\,,\\
\end{gathered}
\end{equation}
by backpropagating the gradient flow through the derivable occlusion model. 
For most occlusions where transparency depends on the model, we in fact consider the blending mask $\alpha = \alpha(w, z)$.
Note that it is required to freeze the discriminator otherwise we would lose any feedback capabilities on the images of the target domain.
For simplicity in Fig.~\ref{fig:advattack}, we omit the generator $G'$ during parameter estimation since $G'(x) = x$.
Training until convergence, we extract the optimal parameter set $\tilde{w}$. 
In Sec.~\ref{sec:exp-raindrops-parameters} we evaluate our parameters estimation on synthetic and real data.

Alternately, $\tilde{w}$ could also be tuned manually but at the cost of menial labor and obvious approximation. Still, one may note than an inaccurate estimation of $\tilde{w}$ would lead to a poor disentanglement of $P_{Y_S}$ and $P_{Y_O}$.

\subsection{Disentanglement guidance} \label{sec:domainbias}
We highlight now an easy pitfall in the disentangled GAN training, since an unwanted optimum is reached if the generator simply adds occlusions.
Indeed because occlusions are visually simple and constitute a strong discriminative signal for the discriminator, it may be easier for the generator to entangle occlusions rather than to learn the underlying scene mapping. 
Specifically, occlusions will be entangled where source and target differ the least, since it is an easy way to minimize $L_{\text{gen}}$ as even with a perfect i2i \textit{there} the discriminator will provide relatively uncertain feedback.
For example, we noticed that drops were entangled over trees or buildings as both exhibit little visual differences in the clear and rainy domains.

To avoid such undesirable behavior, we spatially guide the disentanglement to prevent the i2i task from entangling occlusions in the scene representation. 
The so-called \textit{disentanglement guide} is computed through the estimation of the domains gap database-wide. 
Specifically, we use GradCAM~\cite{selvaraju2017grad} which relies on gradient flow through the discriminator to identify which regions contribute to the \textit{fake} classification, and thus exhibit a large domain gap. 
Similar to the parameter estimation (Sec.~\ref{sec:advmodel}), we pretrain a simple i2i baseline and exploit the discriminator. 
To preserve resolution, we upscale and average the response of GradCAM for each discriminator layer and further average responses over the dataset\footnote{Note that averaging through the dataset implies similar image aspects and viewpoints. Image-wise guidance could be envisaged at the cost of less reliable guidance.}. Formally, using LSGAN we extract \textit{Disentanglement Guidance (DG)}
\begin{equation}
DG = \mathbb{E}_{x~\sim P_X(x)}[\mathbb{E}_{l \in L}[\text{GradCAM}_l(D(x))]\,,
\label{eq:method-dg}
\end{equation}
with $L$ being the discriminator layers. During training of our method, the guide serves to inject occlusions only where domain gaps are low, that is where $DG < \beta$, with $\beta \in [0, 1]$ a hyperparameter. 
While this may seem counter-intuitive, explicitly injecting drops in low domain shift areas 
mimics the GAN intended behavior lowering the domain shift with drops. This logically prevents entanglement phenomena since they are simulated by the injection of occlusions. We refer to Fig.~\ref{fig:ablation-beta} in the ablation study for a visual understanding of this phenomenon.

\section{Experiments}
\label{sec:exp}

We validate the performances of our method on various real occlusions, leveraging recent real datasets such as nuScenes~\cite{caesar2019nuscenes}, RobotCar~\cite{porav2019can}, Cityscapes~\cite{cordts2016cityscapes} or WoodScape~\cite{yogamani2019woodscape}, and synthetic data such as Synthia~\cite{ros2016synthia}.
Our most comprehensive results focus on the harder raindrops occlusion~(Sec.~\ref{sec:exp-raindrops}), but we also extend to soil/dirt and other general occlusions such as watermark, fence, etc. (Sec.~\ref{sec:exp-general}). 
For each type of occlusion we detail the model used and report qualitative and quantitative results against recent works: DRIT~\cite{lee2019drit++}, U-GAT-IT~\cite{kim2019u}, AttentionGAN~\cite{tang2019attention}, CycleGAN~\cite{zhu2017unpaired}, and MUNIT~\cite{huang2018multimodal}.
Because the literature does not account for disentanglement, we report both the disentangled underlying domain (\textit{Ours disentangled}) and the disentangled domain \textit{with} injection of \textit{target} occlusions (\textit{Ours target}). Note that while still images are already significantly better with our method, the full extent is better sensed on the supplementary video as disentangling domains implicitly enforces temporal consistency.

\subsection{Training setup}
\label{sec:meth-training}
Our method is trained in a three stages unsupervised fashion, with the only prior that the occlusion model is known (e.g. drop, dirt, watermark, etc.).
First, we train an i2i baseline to learn the entangled $\text{source}\mapsto{}\text{target}$ and extract $D^{\text{ent}}$. Second, the occlusion model parameters are regressed as in Sec.~\ref{sec:advmodel} and DG is estimated with the same pre-trained discriminator following Sec.~\ref{sec:domainbias}. While being not mandatory for disentanglement, DG often improves visual results. Third, the disentangled pipeline described in Sec.~\ref{sec:disentanglement} is trained from scratch injecting the occlusion model only where the Disentanglement Guidance allows it. We refer to the supplementary for more details.\\
We use MUNIT~\cite{huang2018multimodal} for its multi-modal capacity and train with LSGAN~\cite{mao2017least}. Occlusion models are implemented in a differentiable manner with Kornia~\cite{riba2019kornia}.

\subsection{Raindrops}
\label{sec:exp-raindrops}

We now evaluate our method on the complex task of raindrops disentanglement when learning the i2i $\text{clear}\mapsto{}\text{rain}$ task. 
Because of their refractive and semi-transparent appearance, raindrops occlusions are fairly complex. 

\subsubsection{Occlusion model.}
\begin{figure}
	\centering
	\includegraphics[width=0.7\textwidth]{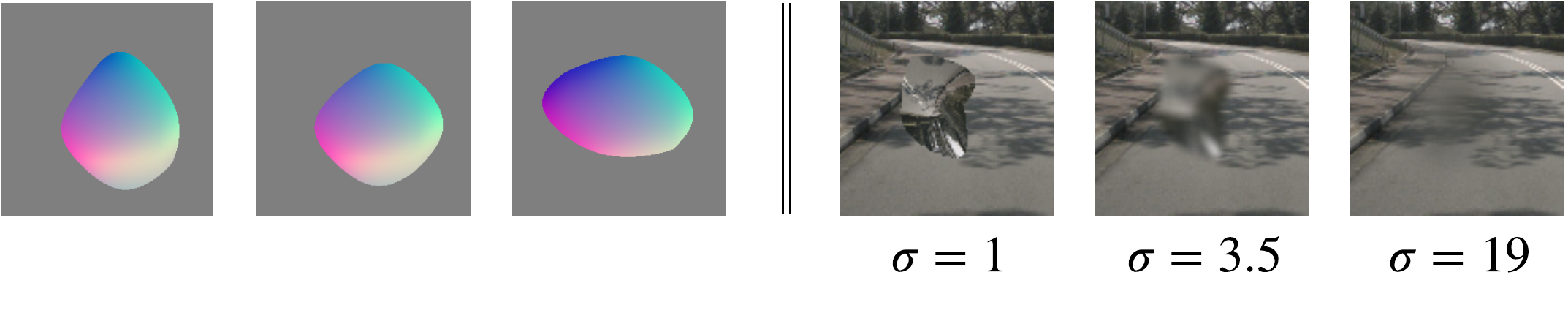}
	\caption{Raindrop occlusion model. Left are schematic views of our model where the shape is modeled as trigonometric functions and photometry as displacement maps (cf. Eq.~\ref{eq:exp-raindrops-uv}), encoded here as RGB. Right demonstrates our ability to handle the high variability of drops appearance with a different focus ($\sigma$).}
	\label{fig:drops}
\end{figure}

To model raindrops, we use the recent model of Alletto \textit{et al.}~\cite{alletto2019adherent} which provides a good realism/simplicity trade-off.
Following~\cite{alletto2019adherent}, we approximate drop shapes with simple trigonometric functions and add random noise to increase variability as in~\cite{shadertoy}. 
The photometry of drops is approached with a displacement map $(U, V)$ encoding the 2D coordinate mapping in the target image, such that drop at $(u,v)$ with thickness $\rho$ has its pixel $(u_i, v_i)$ mapped to
\begin{equation}
    \big(u + \text{U}(u_i,v_i)\cdot\rho, v + \text{V}(u_i,v_i)\cdot\rho\big)\,.
    \label{eq:exp-raindrops-uv}
\end{equation}
Intuitively, this approximates light refractive properties of raindrops. Technically, $(\text{U},\text{V},\rho)$ is conveniently encoded as a 3-channels image. We refer to~\cite{alletto2019adherent} for details.\\
We also account for the imaging focus, as it has been highlighted that drops with different focus have dramatically different appearance~\cite{halimeh2009raindrop,cord2011towards,alletto2019adherent}. We approximate focus blur with a Gaussian point spread function~\cite{pentland1987new} which variance $\sigma$ is learned, thus $w = \{\sigma\}$. I.e., our method handles drops occlusions with any type of focus.\\
During training, drops are uniformly distributed in the image space, with size being a hyperparameter which we study later, and defocus $\sigma$ is regressed with our parameters estimation~(Sec.~\ref{sec:advmodel}). During inference, drops are generated at random position with $p_{\text{r}}$ probability, which somehow controls the rain intensity.\\
Fig.~\ref{fig:drops} illustrates our drop occlusion model with variable shapes and focus blur.

\subsubsection{Datasets.}
We evaluate using 2 recent datasets providing clear/rain images.\\
\noindent{}\textit{nuScenes}~\cite{caesar2019nuscenes} is an urban driving dataset recorded in the US and Singapore with coarse frame-wise weather annotation. Using the latter, we split the validation into clear/rain and obtain 114251/29463 for training and 25798/5637 for testing.

\noindent{}\textit{RobotCar}~\cite{porav2019can} provides pairs of clear/rain images acquired with binocular specialized hardware where one camera is continuously sprayed with water. The clear images are warped to the rainy image space using calibration data, and we use a clear/rain split of 3816/3816 for training and 1000/1000 for validation.

\subsubsection{Qualitative evaluation.}
\begin{figure}
	\centering
	\resizebox{\linewidth}{!}{
	\setlength{\tabcolsep}{0.003\linewidth}
	\tiny
	\begin{tabular}{c c c c c c c}
			& \multirow{1}{*}[0.35cm]{\rotatebox{90}{Target}}
			& \includegraphics[width=10em, valign=m]{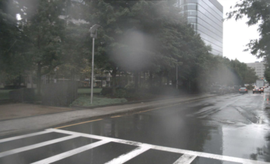}
			& \includegraphics[width=10em, valign=m]{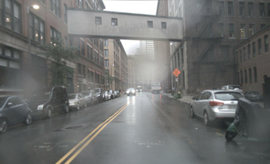}
			& \includegraphics[width=10em, valign=m]{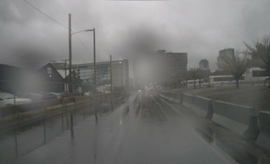}
			& \includegraphics[width=10em, valign=m]{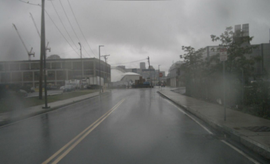}
			& \includegraphics[width=10em, valign=m]{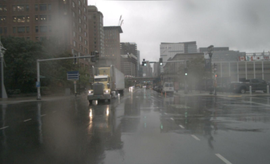}%
			\\
			\midrule
			& \multirow{1}{*}[0.35cm]{\rotatebox{90}{Source}}
			& \includegraphics[width=10em, valign=m]{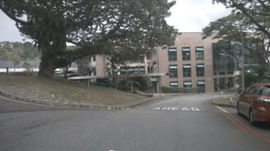}
			& \includegraphics[width=10em, valign=m]{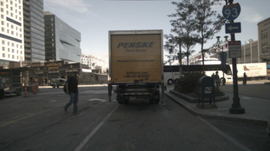}
			& \includegraphics[width=10em, valign=m]{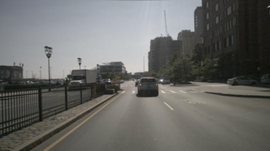}
			& \includegraphics[width=10em, valign=m]{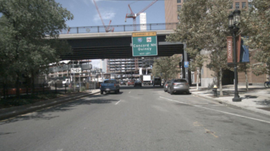}
			& \includegraphics[width=10em, valign=m]{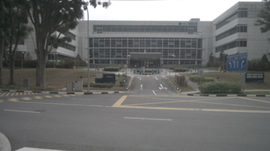}%
			\\
			\multirow{1}{*}[0.15cm]{\rotatebox{90}{\cite{zhu2017unpaired}}} & \multirow{1}{*}[0.45cm]{\rotatebox{90}{CycleGAN}}
			& \includegraphics[width=10em, valign=m]{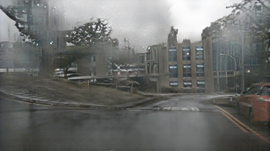}
			& \includegraphics[width=10em, valign=m]{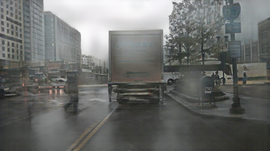}
			& \includegraphics[width=10em, valign=m]{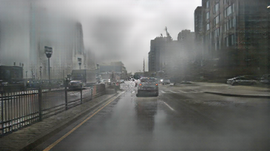}
			& \includegraphics[width=10em, valign=m]{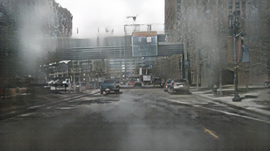}
			& \includegraphics[width=10em, valign=m]{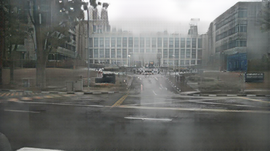}%
			\\
			
			\multirow{1}{*}[0.15cm]{\rotatebox{90}{\cite{tang2019attention}}} & \multirow{1}{*}[0.53cm]{\rotatebox{90}{Attent.GAN}}
			& \includegraphics[width=10em, valign=m]{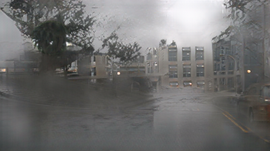}
			& \includegraphics[width=10em, valign=m]{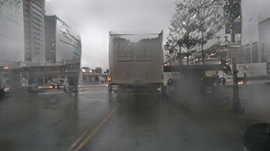}
			& \includegraphics[width=10em, valign=m]{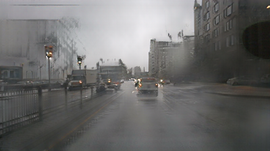}
			& \includegraphics[width=10em, valign=m]{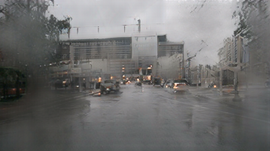}
			& \includegraphics[width=10em, valign=m]{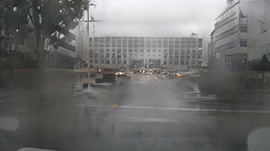}%
			\\
			
			\multirow{1}{*}[0.15cm]{\rotatebox{90}{\cite{kim2019u}}} & \multirow{1}{*}[0.42cm]{\rotatebox{90}{U-GAT-IT}}
			& \includegraphics[width=10em, valign=m]{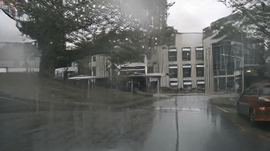}
			& \includegraphics[width=10em, valign=m]{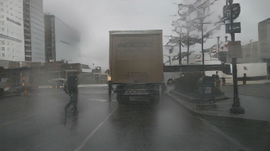}
			& \includegraphics[width=10em, valign=m]{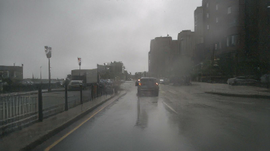}
			& \includegraphics[width=10em, valign=m]{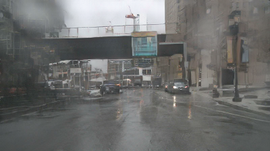}
			& \includegraphics[width=10em, valign=m]{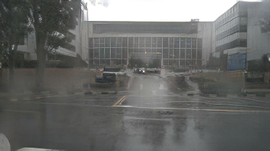}%
			\\
			\multirow{1}{*}[0.15cm]{\rotatebox{90}{\cite{lee2019drit++}}} & \multirow{1}{*}[0.24cm]{\rotatebox{90}{DRIT}}
			& \includegraphics[width=10em, valign=m]{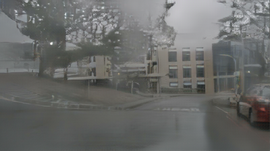}
			& \includegraphics[width=10em, valign=m]{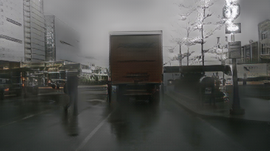}
			& \includegraphics[width=10em, valign=m]{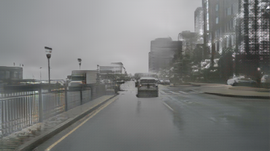}
			& \includegraphics[width=10em, valign=m]{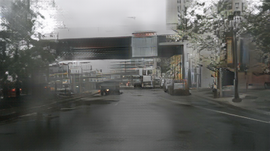}
			& \includegraphics[width=10em, valign=m]{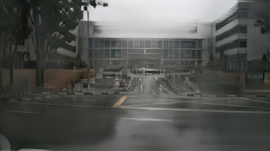}%
			\\
			
			\multirow{1}{*}[0.15cm]{\rotatebox{90}{\cite{huang2018multimodal}}} & \multirow{1}{*}[0.3cm]{\rotatebox{90}{MUNIT}}
			& \includegraphics[width=10em, valign=m]{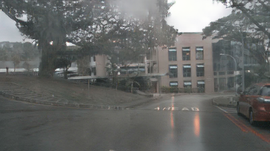}
			& \includegraphics[width=10em, valign=m]{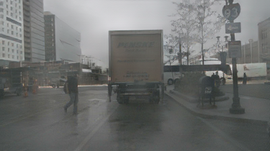}
			& \includegraphics[width=10em, valign=m]{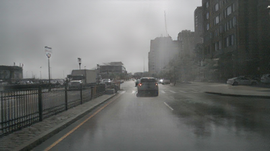}
			& \includegraphics[width=10em, valign=m]{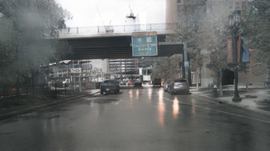}
			& \includegraphics[width=10em, valign=m]{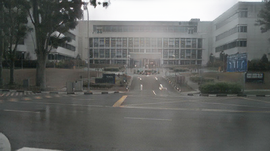}%
			\\
			\midrule
			\multirow{1}{*}[0.2cm]{\rotatebox{90}{Ours}} & \multirow{1}{*}[0.54cm]{\rotatebox{90}{disentangled}}
			& \includegraphics[width=10em, valign=m]{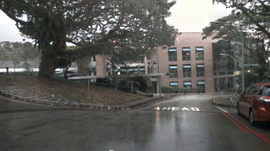}
			& \includegraphics[width=10em, valign=m]{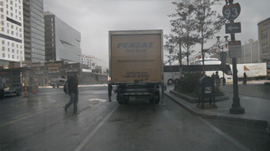}
			& \includegraphics[width=10em, valign=m]{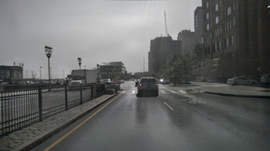}
			& \includegraphics[width=10em, valign=m]{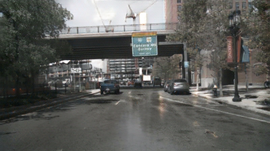}
			& \includegraphics[width=10em, valign=m]{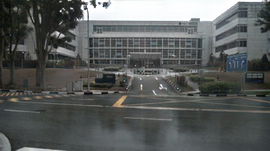}%
			\\
			\multirow{1}{*}[0.2cm]{\rotatebox{90}{Ours}} & \multirow{1}{*}[0.25cm]{\rotatebox{90}{target}}
			& \includegraphics[width=10em, valign=m]{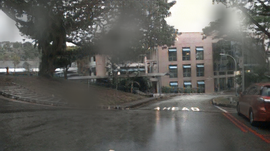}
			& \includegraphics[width=10em, valign=m]{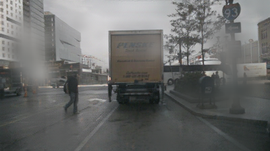}
			& \includegraphics[width=10em, valign=m]{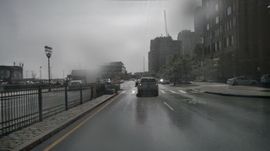}
			& \includegraphics[width=10em, valign=m]{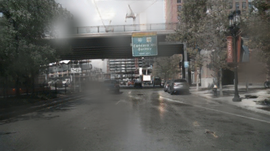}
			& \includegraphics[width=10em, valign=m]{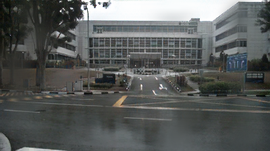}%
			\\
			\multirow{1}{*}[0.2cm]{\rotatebox{90}{Ours}} & \multirow{1}{*}[0.5cm]{\rotatebox{90}{dashcam 1}}
			& \includegraphics[width=10em, valign=m]{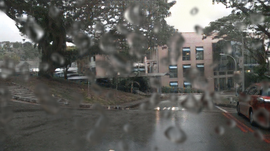}
			& \includegraphics[width=10em, valign=m]{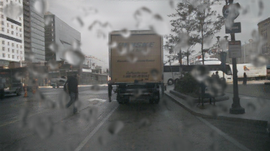}
			& \includegraphics[width=10em, valign=m]{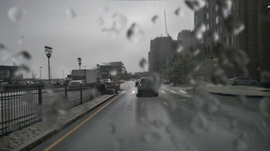}
			& \includegraphics[width=10em, valign=m]{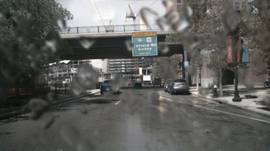}
			& \includegraphics[width=10em, valign=m]{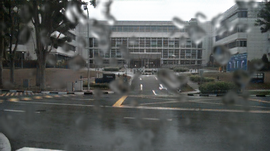}%
			\\
			\multirow{1}{*}[0.2cm]{\rotatebox{90}{Ours}} & \multirow{1}{*}[0.5cm]{\rotatebox{90}{dashcam 2}}
			& \includegraphics[width=10em, valign=m]{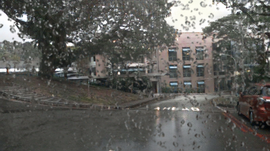}
			& \includegraphics[width=10em, valign=m]{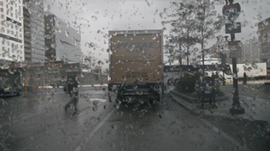}
			& \includegraphics[width=10em, valign=m]{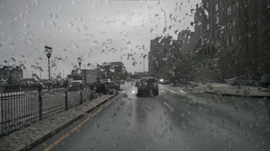}
			& \includegraphics[width=10em, valign=m]{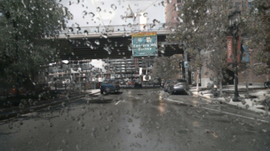}
			& \includegraphics[width=10em, valign=m]{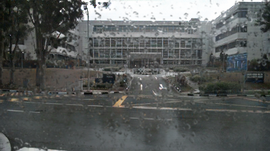}
		\end{tabular}}
		\caption{Qualitative comparison against recent baselines on the $\text{clear}\mapsto{}\text{rain}$ task with drops occlusions. Target samples are displayed in the first row for reference. Other rows show the source image (2nd row) and all its subsequent translations below.
		Our method efficiently disentangled drops occlusion from the scene (row \textit{Ours disentangled}) and subsequently allows the generation of realistic drops matching target style (row \textit{Ours target}) or any arbitrary style (last 2 rows).}\label{fig:qualitative}
\end{figure}

Outputs of the $\text{clear}\mapsto{}\text{rain}$ i2i task are shown in Fig.~\ref{fig:qualitative} against the above cited \cite{lee2019drit++,kim2019u,tang2019attention,zhu2017unpaired,huang2018multimodal}.
At first sight, it is evident that drops are entangled in other methods, which is expected as they are \textit{not} taught to disentangle drops.
To allow fair comparison we thus provide our disentangled estimation (\textit{Ours disentangled}) but also add drops occlusions to it, modeled with the physical parameters $\tilde{w}$ estimated from target domain (\textit{Ours target}).

Looking at \textit{Ours disentangled}, the i2i successfully learned the appearance of a rainy scene (e.g. reflections or sky) with sharp pleasant translation to rain, without any drops.
Other methods noticeably entangle drops, often at fixed positions to avoid learning i2i translation. This is easily noticed in the 4th column where all methods generated drops on the leftmost tree.
Conversely, we benefit from our disentangled representation to render scenes with drops occlusions (row \textit{Ours target}) fairly matching the appearance of the target domain (1st row), subsequently demonstrating the efficiency of our adversarial parameter estimation.

What is more, we inject drops with different sets of parameters $\{w, z\}$ arbitrary mimicking dashcam sequences (Fig.~\ref{fig:qualitative}, last 2 rows). 
The quality of the dashcam translations, despite the absence of similar data during training, proves the benefit of disentanglement and the adequacy of the occlusion model. Note that with any set of parameters, our occlusion (last 3 rows) respect the refractive properties of raindrops showing the scene up-side-down in each drop, while other baselines simply model white and blurry occlusions.

\subsubsection{Quantitative evaluation.} 
\paragraph{GAN metrics.}
Tab.~\ref{table:ganmetrics} reports metrics on the nuScenes $\text{clear}\mapsto{}\text{rain}$ task. Each metric encompasses different meanings: Inception Score (IS)~\cite{salimans2016improved} evaluates quality/diversity against target, LPIPS distance~\cite{zhang2018unreasonable} evaluates translation diversity thus avoiding mode-collapse, and Conditional Inception Score~\cite{huang2018multimodal} single-image translations diversity for multi-modal baselines.
Note that we evaluate against our disentangled + target drops occlusion \textit{Ours target}, since baselines are neither supposed to disentangle the occlusion layer nor to generate different kinds of drop. 
On all metrics, our method outperforms the state of the art by a comfortable margin.
This is easily ascribable to our output being both more realistic, since we are evaluating with drops with the physical parameters extracted from target dataset, and more variable, since we do not suffer from entanglement phenomena that greatly limit the drops visual stochasticity. 
This is also evident when comparing against \cite{huang2018multimodal} which we use as the backbone in our framework.

\noindent{}Technically, IS is computed over the whole validation set, CIS on 100 translations of 100 random images (as in~\cite{huang2018multimodal}), and LPIPS on 1900 random pairs of 100 translations. The InceptionV3 for IS/CIS was finetuned on source/target as \cite{huang2018multimodal}.

\begin{table}[t]
	\centering
	\begin{subfigure}{.45\linewidth}
		\centering
		\scriptsize
    	\setlength{\tabcolsep}{0.007\linewidth}
		\begin{tabular}{cccc}
    		\toprule
			\textbf{Network} & \textbf{IS}$\uparrow$ &  \textbf{LPIPS}$\uparrow$ & \textbf{CIS}$\uparrow$\\
 			\hline
			CycleGAN \cite{zhu2017unpaired} & 1.151 & 0.473 & - \\
			AttentionGAN \cite{tang2019attention} & 1.406 & 0.464 & - \\
			U-GAT-IT \cite{kim2019u} & 1.038 & 0.489 & - \\
 			\hline %
			DRIT \cite{lee2019drit++} & 1.189 & 0.492 & 1.120 \\
			MUNIT \cite{huang2018multimodal} & 1.211 & 0.495 & 1.030\\
			\hline %
			Ours target & \textbf{1.532}  & \textbf{0.515} & \textbf{1.148} \\
    		\bottomrule
		\end{tabular}
		\caption{GAN metrics}\label{table:ganmetrics}
	\end{subfigure}%
    \begin{subfigure}{.45\linewidth}
		\centering
		\scriptsize
    	\setlength{\tabcolsep}{0.007\linewidth}
        \begin{tabular}{cc}
    		\toprule
			\textbf{Method} & \textbf{AP$\uparrow$} \\
			\hline Original (from~\cite{halder2019physics}) & 18.7  \\			
			\makecell{\tiny{Finetuned w/}\\Halder \textit{et al.}~\cite{halder2019physics}} & 25.6  \\			
			\hline \makecell{\tiny{Finetuned w/}\\Ours target} & \textbf{27.7} \\
			\bottomrule
		\end{tabular}
		\caption{Semantic segmentation}\label{table:semantic}
    \end{subfigure}
	
	\caption{Quantitative evaluation of $\text{clear}\mapsto{}\text{rain}$ effectiveness on nuScenes~\cite{caesar2019nuscenes}  (for all higher is better). (\protect\subref{table:ganmetrics}) shows GAN metrics of ours i2i translation with target drop inclusion (i.e. \textit{Ours Target}) against i2i baselines. Our method outperforms literature on all metrics which is imputed to the variability and realism that come with the disentanglement. Note that CIS is multi-modal. (\protect\subref{table:semantic}) Evaluation of the Average Precision (AP) of semantic segmentation when finetuning PSPNet~\cite{zhao2017pyramid} and evaluating on a subset of nuScenes with semantic labels from~\cite{halder2019physics}.}
\end{table}

\paragraph{Semantic segmentation.}
Because GAN metrics are reportedly noisy~\cite{zhang2018unreasonable} we aim at providing a different perspective for quantitative evaluation, and thus measure the usefulness of our translated images for semantic segmentation.
To that aim, following the practice of Halder et al.~\cite{halder2019physics} we use \textit{Ours target} (i.e. trained on rainy nuScenes) to infer a rainy version of the popular Cityscapes~\cite{cordts2016cityscapes} dataset and use it to finetune PSPNet~\cite{zhao2017pyramid}.
The evaluation on the small subset of 25 semantic labeled images of rainy nuScenes provided by~\cite{halder2019physics} is reported in Tab.~\ref{table:semantic}. It showcases finetuning with our rainy images is better than with~\cite{halder2019physics}, which uses physics-based rendering to generate rain.
Note that both finetune \textit{Original} weights, and that the low numbers results of the fairly large Cityscapes-nuScenes gap (recall that nuScenes has no semantic labels to train on).

\subsubsection{Parameter estimation.}
\label{sec:exp-raindrops-parameters}
\begin{figure}[t]
    \centering
	\begin{subfigure}[b]{0.45\textwidth}
    	\centering
    	\tiny
    	\setlength{\tabcolsep}{0.003\linewidth}
    	\renewcommand{\arraystretch}{0.3}
    	\begin{tabular}{cccc}
    	     \includegraphics[width=.24\textwidth]{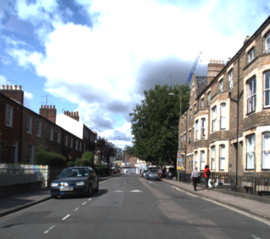} & \includegraphics[width=.24\textwidth]{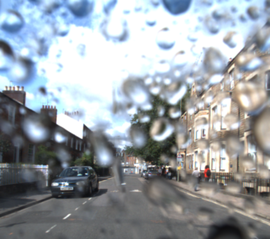} &
    	     \includegraphics[width=.24\textwidth]{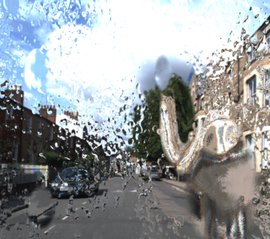} & \includegraphics[width=.24\textwidth]{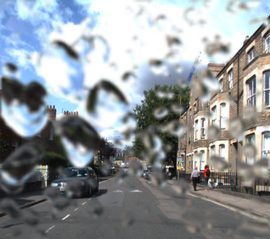} \\
    	     \\
    	     Source & Target & \cite{porav2019can} & Ours target
    	\end{tabular}
		\caption{Sample images}\label{fig:porav-qualitative}
	\end{subfigure}
	\begin{subfigure}[b]{0.28\textwidth}
        \tiny
    	\setlength{\tabcolsep}{0.003\linewidth}
        \begin{tabular}{ccc}
            \toprule
    		\textbf{Method} & \textbf{FID}$\downarrow$ & \textbf{LPIPS}$\downarrow$ \\
    		\hline Porav et al. \cite{porav2019can} & 207.34 & 0.53  \\			
    		Ours target & \textbf{135.32} & \textbf{0.44} \\
    		\bottomrule
    	\end{tabular}
    	\caption{Benchmark on \cite{porav2019can}}
    	\label{fig:porav-dist}
	\end{subfigure}
	\begin{subfigure}[b]{0.25\textwidth}
    	\centering
	    \includegraphics[width = 1.0\textwidth]{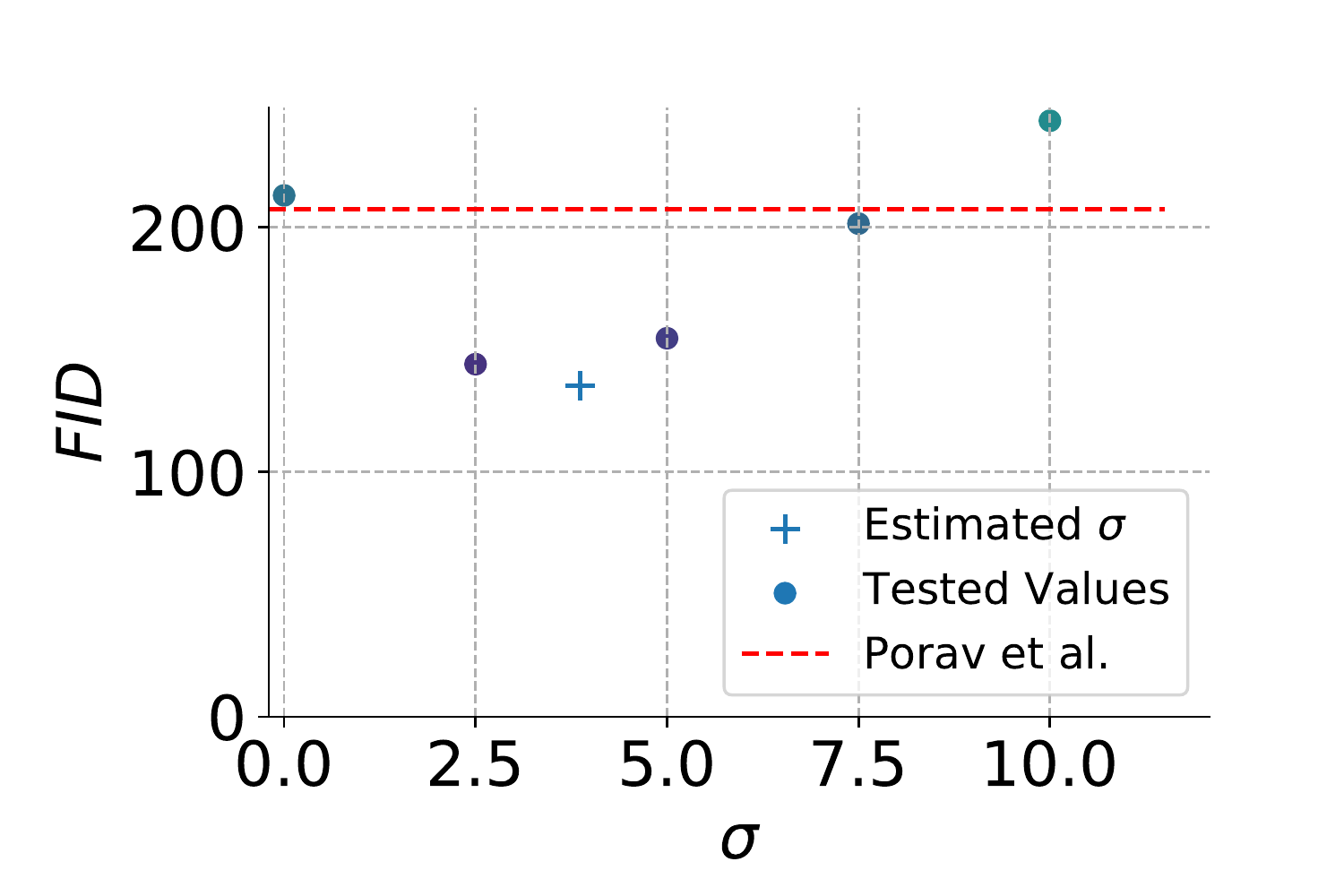}
    	\caption{FID}
    	\label{fig:porav-fid}
	\end{subfigure} %
	\caption{Parameter estimation using real clear/rainy pairs of images from RobotCar~\cite{porav2019can}. Visually, \textit{Ours target} is fairly closer to the \textit{Target} sample regardless of drops position/size (\protect\subref{fig:porav-qualitative}), while quantitatively lower FID and LPIPS distance is obtained (\protect\subref{fig:porav-dist}). With FID measures at different defocus sigma in (\protect\subref{fig:porav-fid}), we demonstrate our estimated parameters ($\sigma=3.87$) successfully led to the best parameters.}\label{fig:porav}
\end{figure}

We verify the validity of our parameter estimation strategy (Sec.~\ref{sec:advmodel}) using the RobotCar dataset, which provides real \textit{clear/rain} pairs of images. As the viewpoints are warped together (cf. Datasets details above), there is no underlying domain shift in the clear/rain images so we set $G(x) = x$ and directly train on discriminator to regress physical parameters (we get $\sigma=3.87$) and render rain with it on clear images.
We can then measure the distance of the translated and real rainy images, with FID and LPIPS distances reported in Fig.~\ref{fig:porav-dist}. Unlike before, LPIPS measures distance (not diversity) so lower is better. For both we significantly outperform \cite{porav2019can}, which is visually interpretable in Fig.~\ref{fig:porav-qualitative}, where drops rendered with our parameter estimation looks more similar to \textit{Target} than those of \cite{porav2019can} (regardless of their size/position).

To further assess the accuracy of our estimation, we plot in Fig.~\ref{fig:porav-fid} the FID for different defocus blurs ($\sigma \in \{0.0, 2.5, 5.0, 7.5, 10\}$). It shows our estimated defocus ($\sigma=3.87$) leads to the minimum FID of all tested values, further demonstrating the accuracy of our adversarial estimation.
We quantify the precision by training on clear images with synthetic injection of drops having $\sigma \in [5,25]$, and we measured an average error of 1.02\% (std. 1.95\%).

\subsection{Extension to other occlusion models} To showcase the generality of our method, we demonstrate its performance on two generally encountered types of occlusions: Dirt and General occlusion.

\subsubsection{Dirt.}
\label{sec:exp-soil}
We rely on the recent WoodScape dataset~\cite{yogamani2019woodscape} and a simple occlusion model to learn the disentangled representation.

\noindent{}\textit{Datasets.} WoodScape~\cite{yogamani2019woodscape} provides a large amount of driving fish-eye images, and comes with metadata indicating the presence of dirt/soil\footnote{Note that WoodScape provides soiling mask which we do \textit{not} use.}. Different from rain sequences having rainy scenes+drops occlusions, apart from soiling there isn't any domain shift in the clean/dirt images provided. Hence, to study disentanglement we introduce an additional shift by converting clean images to grayscale and refer to them as \textit{clean\_gray}. We train our method on non-paired clean\_gray/dirt images with 5117/4873 for training and 500/500 for validation.

\noindent{}\textit{Occlusion model.} To generate synthetic soiling, we use a modified version of our drop model with random trigonometric functions and larger varying sizes. Displacement maps are not used since we consider dirt to be opaque and randomly brownish, with apparent semi-transparency only as a result of the high defocus. As for drops, the defocus $\sigma$ is regressed so again $w = \{\sigma\}$.

\noindent{}\textit{Performance.} Fig.~\ref{fig:dirt} (left) shows sample results where the task consists of disentangling the color characteristics from the dirt occlusion (since color is only in the dirt data).
Comparing to MUNIT~\cite{huang2018multimodal}, \textit{Ours disentangled} successfully learned color without dirt entanglement, while \cite{huang2018multimodal} failed to learn accurate colorization due to entanglement. Performances are validated quantitatively in Tab.~\ref{tab:gan-metrics-others}-\ref{tab:colorization}.

\subsubsection{General occlusions (synthetic).}
\label{sec:exp-general}
\begin{figure}[t]
	\begin{subfigure}[t]{0.65\textwidth}
	\resizebox{1.0\textwidth}{!}{
	\setlength{\tabcolsep}{0.002\linewidth}
	\scriptsize
    \begin{tabular}{c c c c c c}
            &&\multicolumn{2}{c}{\footnotesize{}\textbf{Dirt~\cite{yogamani2019woodscape}}}&\multicolumn{1}{c}{\footnotesize{}\textbf{WMK~\cite{ros2016synthia}}}&\multicolumn{1}{c}{\footnotesize{}\textbf{Fence~\cite{ros2016synthia}}}\\
            \cmidrule[1pt](){3-4}\cmidrule[1pt](l){5-5}\cmidrule[1pt](l){6-6}
			\rotatebox{90}{\textbf{}} 
			\multirow{1}{*}[0.2cm]{\rotatebox{90}{}} & \multirow{1}{*}[0.4cm]{\rotatebox{90}{Target}}
			& \includegraphics[width=9em, valign=m]{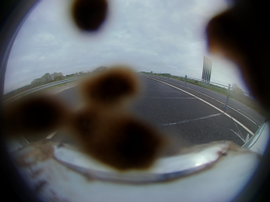}
			& \includegraphics[width=9em, valign=m]{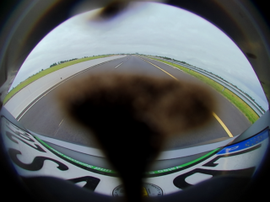}
			& %
			\includegraphics[width=12em, valign=m]{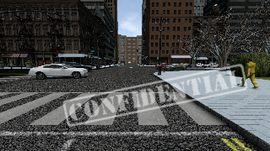}
			& \includegraphics[width=12em, valign=m]{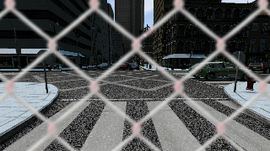}%
			\\ 
			
			\midrule
			\multirow{1}{*}[0.2cm]{\rotatebox{90}{}} & \multirow{1}{*}[0.4cm]{\rotatebox{90}{Source}}
			& \includegraphics[width=9em, valign=m]{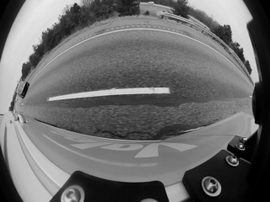}
			& \includegraphics[width=9em, valign=m]{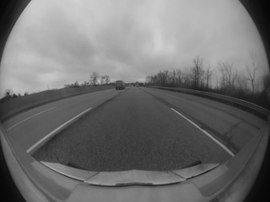}
			& %
			\includegraphics[width=12em, valign=m]{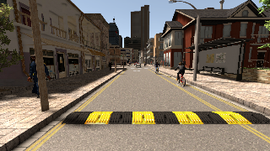}
			& \includegraphics[width=12em, valign=m]{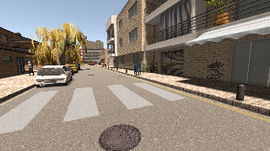}%
			\\ 
			
    		& \multirow{1}{*}[0.7cm]{\rotatebox{90}{MUNIT~\cite{huang2018multimodal}}}
			& \includegraphics[width=9em, valign=m]{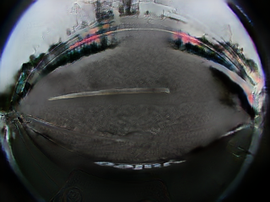}
			& \includegraphics[width=9em, valign=m]{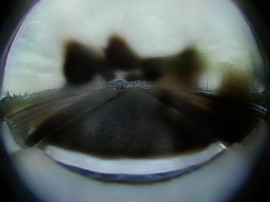}
			& %
			\includegraphics[width=12em, valign=m]{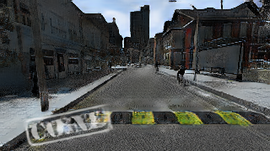}
			& \includegraphics[width=12em, valign=m]{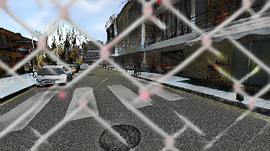}%
			\\ 
			
			\midrule
			\multirow{1}{*}[0.2cm]{\rotatebox{90}{Ours}} & \multirow{1}{*}[0.65cm]{\rotatebox{90}{Disentangled}}
			& \includegraphics[width=9em, valign=m]{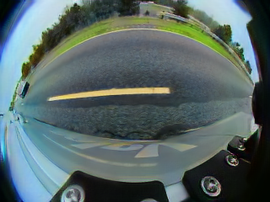}
			& \includegraphics[width=9em, valign=m]{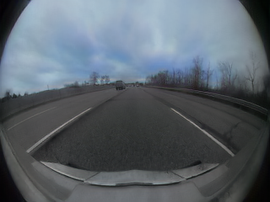}
			& %
			\includegraphics[width=12em, valign=m]{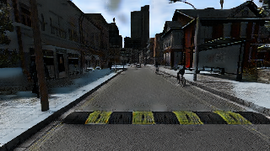}
			& \includegraphics[width=12em, valign=m]{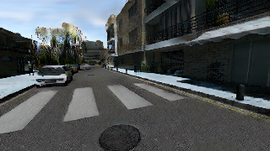}%
			\\ 
			
		\end{tabular}}
		\caption{Qualitative results}\label{fig:qualitative-other}\label{fig:dirt}\label{fig:toy}
    \end{subfigure}\hspace{0.01\textwidth}\begin{subfigure}[t]{.34\textwidth}
    	\begin{subfigure}[t!]{\textwidth}
        \tiny
    	\setlength{\tabcolsep}{0.007\linewidth}
            \begin{tabular}{ccccc}
            \toprule
            \textbf{Model}                  & \textbf{Network}          & \textbf{IS$\uparrow$}    & \textbf{LPIPS$\uparrow$} & \textbf{CIS$\uparrow$}   \\\hline
            \multirow{2}{*}{Dirt}      & MUNIT~\cite{huang2018multimodal} & 1.06 & \textbf{0.66} & 1.08 \\
                                       & Ours target             & \textbf{1.26} & 0.59 & \textbf{1.15} \\ \hline
            \multirow{2}{*}{Fence}     & MUNIT~\cite{huang2018multimodal} & 1.26 & \textbf{0.55} & 1.11 \\
                                       & Ours target           & \textbf{1.31} & 0.54 & \textbf{1.19} \\\hline
            \multirow{2}{*}{WMK} & MUNIT~\cite{huang2018multimodal} & 1.17 & \textbf{0.57} & 1.01 \\
                                       & Ours target            & \textbf{1.19} & 0.55 & \textbf{1.02}\\
            \bottomrule
        
            \end{tabular}        \caption{GAN metrics}
        \label{tab:gan-metrics-others}

	    \end{subfigure}%
	    \hfill
	    \par\smallskip\smallskip\smallskip
    	\begin{subfigure}[t]{\textwidth}
            \tiny
        	\setlength{\tabcolsep}{0.007\linewidth}
        	\begin{tabular}{cccc}
            \toprule
            \textbf{Model} & \textbf{Network} & \textbf{SSIM$\uparrow$} & \textbf{PSNR$\uparrow$} \\ \hline
            \multirow{2}{*}{Dirt}      & MUNIT~\cite{huang2018multimodal}  & 0.41 & 13.40 \\
            & Ours disent. & \textbf{0.76} & \textbf{20.23} \\
            \bottomrule
            \end{tabular}\caption{Colorization metrics}\label{tab:colorization}
        \end{subfigure}
        \vfil
    \end{subfigure}
	\caption{Various occlusions disentanglement. We seek to learn disentangled representation of $\text{clean\_gray}\mapsto{}\text{clean\_color}$ on WoodScape~\cite{yogamani2019woodscape} (real) and $\text{clear}\mapsto{}\text{snow}$ on Synthia~\cite{ros2016synthia} (synthetic). For all, MUNIT~\cite{huang2018multimodal} partly entangles occlusions in the translation, often occluding hard-to-translate areas, while our method learned correctly the color mapping and the snow mapping despite complex occlusions (\ref{fig:qualitative-other}). Quantitative evaluation with GAN metrics  (\ref{tab:gan-metrics-others}) confirms the increase in image quality for all occlusions models and with colorization metrics for dirt (\ref{tab:colorization}) exploiting our unpaired disentanglement framework.}
\end{figure}

In Fig.~\ref{fig:toy} (right) we also demonstrate the ability to disentangle general occlusion, in the sense of an alpha-blended layer on an image (watermarks, logos, etc.). We used synthetic Synthia~\cite{ros2016synthia} clear/snow data, and augmented only snow either with a "confidential" watermark (WMK) or a fence image, both randomly shifted. Our i2i takes 3634/3739 clear/snow images for training, and 901/947 for validation. The occlusion model is the ground truth composite alpha-blended model, with random translation, and without any regressed parameters (i.e. $w=\emptyset$).
From Fig.~\ref{fig:toy}, our method learned a disentangled representation, while MUNIT~\cite{huang2018multimodal} partially entangled the occlusion model. In tab.~\ref{tab:gan-metrics-others}, CIS/IS confirm the higher quality visual results.

\subsection{Ablation studies}
\label{sec:exp-ablation}

\begin{figure}
	\centering
    \scriptsize
	\begin{subfigure}{.28\linewidth}
        {\renewcommand{\arraystretch}{1.1}\tiny
    	\setlength{\tabcolsep}{0.007\linewidth}
            \begin{tabular}{cccc}
            \toprule
            \textbf{Model}          & \textbf{IS$\uparrow$}    & \textbf{LPIPS$\uparrow$} & \textbf{CIS$\uparrow$}   \\\hline
            N/A (\cite{huang2018multimodal}) & 1.21 & 0.50 & 1.03 \\\hline
            Gaussian & 1.35 & 0.51 & 1.13 \\
            Refract & 1.46 & 0.50 & 1.12 \\
            Ours & \textbf{1.53} & \textbf{0.52} & \textbf{1.15} \\
            \bottomrule
        
            \end{tabular}}
    \caption{Model complexity}
    \label{fig:ablation-modelcomp}
	\end{subfigure}\begin{subfigure}{.72\linewidth}
    {\renewcommand{\arraystretch}{0.2}
	\setlength{\tabcolsep}{0.002\textwidth}
    \begin{tabular}{cccc}
    \includegraphics[width=0.248\textwidth]{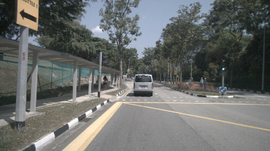}&\includegraphics[width=0.248\textwidth]{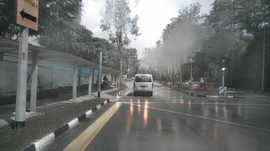}&\includegraphics[width=0.248\textwidth]{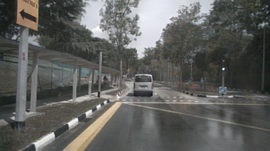}&\includegraphics[width=0.248\textwidth]{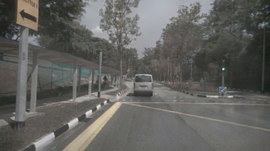}\\
    Source & $\beta = 0$ {{(i.e. \cite{huang2018multimodal})}} & $\beta = 0.75$ {{(ours)}} & $\beta = 1$
    \end{tabular} }
    \caption{Disentanglement guidance}
    \label{fig:ablation-beta}
    \end{subfigure}
	\caption{Ablation of model complexity and disentanglement guidance for the $\text{clear}\mapsto{}\text{rain}$ task on nuScenes. In (\protect\subref{fig:ablation-modelcomp}), our disentanglement performs better than baseline \cite{huang2018multimodal} with all occlusion models. In (\protect\subref{fig:ablation-beta}), studying the influence of $\beta$ we note that without guidance ($\beta = 1$) the translation lacks important rainy features (reflections, glares, etc.) while with appropriate guidance ($\beta=0.75$) it learns correct rainy characteristics without entanglement.
	}
\end{figure}

\paragraph{Model complexity.} We study here how much model complexity impacts disentanglement, with the evaluation on the nuScenes  $\text{clear}\mapsto{}\text{rain}$ task. We compare three decreasingly complex occlusion models: 1)~\textit{Ours}, the raindrop model described in Sec.~\ref{sec:exp-raindrops}; 2)~\textit{Refract}, which is our model without any shape or thickness variability; 3)~\textit{Gaussian}, where drops are modeled as scene-independent Gaussian-shaped occlusion maps following~\cite{gu2009removing}.
From Fig.~\ref{fig:ablation-modelcomp}, while \textit{Ours} has best performance, even simpler models lead to better image translation which we relate to our disentanglement capability.
\noindent{}To also assess that the occlusion model doesn't only play the role of an adversarial attack, we also compare the FID of real RobotCar raindrops (as in Sec.~\ref{sec:exp-raindrops-parameters}) when training with either of the models described in Sec.~\ref{sec:exp-raindrops} and \ref{sec:exp-soil}. The FID measured are $\mathbf{135.32}$ (drop) / $329.17$ (watermark) / $334.76$ (dirt) / $948.71$ (fence). This advocates that \textit{a priori} knowledge of the occlusion type is necessary to achieve good results.

\paragraph{Disentanglement Guidance (DG).} We study the effects of guidance (eq.~\ref{eq:method-dg}) on the nuScenes $\text{clear}\mapsto\text{rain}$ task, by varying the $\beta$ threshold used to inject occlusion where $DG < \beta$. From Fig.~\ref{fig:ablation-beta}, with conservative guidance ($\beta=0$, i.e. no occlusions injected) it behaves similar to MUNIT baseline entangling drops in the translation, while deactivating guidance ($\beta=1$) correctly achieves a disentangled representation but at the cost of losing translation in high domain shifts areas (note the lack of road reflections). Appropriate guidance ($\beta=0.75$) helps learning target characteristics while preserving from entanglement.

\section{Conclusion}
We propose the first unsupervised method for model-based disentanglement in i2i translation, relying on guided injection of occlusions with parameters regressed from target and assuming only prior knowledge of the occlusion model.
Our method outperformed the literature visually and on all tested metrics, and the applicability was shown on various occlusions models (raindrop, dirt, watermark, etc.). 
Our strategy of adversarial parameter estimation copes with drops of any focus, which is of high interest for any outdoor system as demonstrated in the experiments.

\bibliographystyle{splncs04}
\bibliography{egbib}

\begin{thebibliography}{10}
\providecommand{\url}[1]{\texttt{#1}}
\providecommand{\urlprefix}{URL }
\providecommand{\doi}[1]{https://doi.org/#1}

\bibitem{shadertoy}
Rain drops on screen. \url{https://www.shadertoy.com/view/ldSBWW}

\bibitem{alletto2019adherent}
Alletto, S., Carlin, C., Rigazio, L., Ishii, Y., Tsukizawa, S.: Adherent
  raindrop removal with self-supervised attention maps and spatio-temporal
  generative adversarial networks. In: ICCV Workshops (2019)

\bibitem{anoosheh2018combogan}
Anoosheh, A., Agustsson, E., Timofte, R., Van~Gool, L.: Combogan: Unrestrained
  scalability for image domain translation. In: CVPR Workshops (2018)

\bibitem{bi2019deep}
Bi, S., Sunkavalli, K., Perazzi, F., Shechtman, E., Kim, V.G., Ramamoorthi, R.:
  Deep cg2real: Synthetic-to-real translation via image disentanglement. In:
  ICCV (2019)

\bibitem{caesar2019nuscenes}
Caesar, H., Bankiti, V., Lang, A.H., Vora, S., Liong, V.E., Xu, Q., Krishnan,
  A., Pan, Y., Baldan, G., Beijbom, O.: nuscenes: A multimodal dataset for
  autonomous driving. In: CVPR (2020)

\bibitem{cherian2019sem}
Cherian, A., Sullivan, A.: Sem-gan: Semantically-consistent image-to-image
  translation. In: WACV (2019)

\bibitem{choi2018stargan}
Choi, Y., Choi, M., Kim, M., Ha, J.W., Kim, S., Choo, J.: Stargan: Unified
  generative adversarial networks for multi-domain image-to-image translation.
  In: CVPR (2018)

\bibitem{cord2011towards}
Cord, A., Aubert, D.: Towards rain detection through use of in-vehicle
  multipurpose cameras. In: IV (2011)

\bibitem{cordts2016cityscapes}
Cordts, M., Omran, M., Ramos, S., Rehfeld, T., Enzweiler, M., Benenson, R.,
  Franke, U., Roth, S., Schiele, B.: The cityscapes dataset for semantic urban
  scene understanding. In: CVPR (2016)

\bibitem{gu2009removing}
Gu, J., Ramamoorthi, R., Belhumeur, P., Nayar, S.: Removing image artifacts due
  to dirty camera lenses and thin occluders. In: SIGGRAPH Asia (2009)

\bibitem{halder2019physics}
Halder, S.S., Lalonde, J.F., de~Charette, R.: Physics-based rendering for
  improving robustness to rain. In: ICCV (2019)

\bibitem{halimeh2009raindrop}
Halimeh, J.C., Roser, M.: Raindrop detection on car windshields using
  geometric-photometric environment construction and intensity-based
  correlation. In: IV (2009)

\bibitem{hao2019learning}
Hao, Z., You, S., Li, Y., Li, K., Lu, F.: Learning from synthetic
  photorealistic raindrop for single image raindrop removal. In: ICCV Workshops
  (2019)

\bibitem{huang2018multimodal}
Huang, X., Liu, M.Y., Belongie, S., Kautz, J.: Multimodal unsupervised
  image-to-image translation. In: ECCV (2018)

\bibitem{hui2018unsupervised}
Hui, L., Li, X., Chen, J., He, H., Yang, J.: Unsupervised multi-domain image
  translation with domain-specific encoders/decoders. In: ICPR (2018)

\bibitem{isola2017image}
Isola, P., Zhu, J.Y., Zhou, T., Efros, A.A.: Image-to-image translation with
  conditional adversarial networks. In: CVPR (2017)

\bibitem{kim2019u}
Kim, J., Kim, M., Kang, H., Lee, K.: U-gat-it: unsupervised generative
  attentional networks with adaptive layer-instance normalization for
  image-to-image translation. In: ICLR (2020)

\bibitem{lee2019drit++}
Lee, H.Y., Tseng, H.Y., Mao, Q., Huang, J.B., Lu, Y.D., Singh, M., Yang, M.H.:
  Drit++: Diverse image-to-image translation via disentangled representations.
  arXiv preprint arXiv:1905.01270  (2019)

\bibitem{li2018semantic}
Li, P., Liang, X., Jia, D., Xing, E.P.: Semantic-aware grad-gan for
  virtual-to-real urban scene adaption. BMVC  (2018)

\bibitem{liu2017unsupervised}
Liu, M.Y., Breuel, T., Kautz, J.: Unsupervised image-to-image translation
  networks. In: NeurIPS (2017)

\bibitem{liu2019few}
Liu, M.Y., Huang, X., Mallya, A., Karras, T., Aila, T., Lehtinen, J., Kautz,
  J.: Few-shot unsupervised image-to-image translation. In: ICCV (2019)

\bibitem{ma2018gan}
Ma, S., Fu, J., Wen~Chen, C., Mei, T.: Da-gan: Instance-level image translation
  by deep attention generative adversarial networks. In: CVPR (2018)

\bibitem{mao2017least}
Mao, X., Li, Q., Xie, H., Lau, R.Y., Wang, Z., Paul~Smolley, S.: Least squares
  generative adversarial networks. In: ICCV (2017)

\bibitem{mejjati2018unsupervised}
Mejjati, Y.A., Richardt, C., Tompkin, J., Cosker, D., Kim, K.I.: Unsupervised
  attention-guided image-to-image translation. In: NeurIPS (2018)

\bibitem{mo2018instagan}
Mo, S., Cho, M., Shin, J.: Instagan: Instance-aware image-to-image translation.
  ICLR  (2019)

\bibitem{pentland1987new}
Pentland, A.P.: A new sense for depth of field. T-PAMI  (1987)

\bibitem{pizzati2019domain}
Pizzati, F., de~Charette, R., Zaccaria, M., Cerri, P.: Domain bridge for
  unpaired image-to-image translation and unsupervised domain adaptation. In:
  WACV (2020)

\bibitem{porav2019can}
Porav, H., Bruls, T., Newman, P.: I can see clearly now: Image restoration via
  de-raining. In: ICRA (2019)

\bibitem{qu2019enhanced}
Qu, Y., Chen, Y., Huang, J., Xie, Y.: Enhanced pix2pix dehazing network. In:
  CVPR (2019)

\bibitem{ramirez2018exploiting}
Ramirez, P.Z., Tonioni, A., Di~Stefano, L.: Exploiting semantics in adversarial
  training for image-level domain adaptation. In: IPAS (2018)

\bibitem{riba2019kornia}
Riba, E., Mishkin, D., Ponsa, D., Rublee, E., Bradski, G.: Kornia: an open
  source differentiable computer vision library for pytorch. In: WACV (2020)

\bibitem{romero2019smit}
Romero, A., Arbel{\'a}ez, P., Van~Gool, L., Timofte, R.: Smit: Stochastic
  multi-label image-to-image translation. In: ICCV Workshops (2019)

\bibitem{ros2016synthia}
Ros, G., Sellart, L., Materzynska, J., Vazquez, D., Lopez, A.M.: The synthia
  dataset: A large collection of synthetic images for semantic segmentation of
  urban scenes. In: CVPR (2016)

\bibitem{roser2009video}
Roser, M., Geiger, A.: Video-based raindrop detection for improved image
  registration. In: ICCV Workshops (2009)

\bibitem{roser2010realistic}
Roser, M., Kurz, J., Geiger, A.: Realistic modeling of water droplets for
  monocular adherent raindrop recognition using bezier curves. In: ACCV (2010)

\bibitem{salimans2016improved}
Salimans, T., Goodfellow, I., Zaremba, W., Cheung, V., Radford, A., Chen, X.:
  Improved techniques for training gans. In: NeurIPS (2016)

\bibitem{selvaraju2017grad}
Selvaraju, R.R., Cogswell, M., Das, A., Vedantam, R., Parikh, D., Batra, D.:
  Grad-cam: Visual explanations from deep networks via gradient-based
  localization. In: ICCV (2017)

\bibitem{shen2019towards}
Shen, Z., Huang, M., Shi, J., Xue, X., Huang, T.S.: Towards instance-level
  image-to-image translation. In: CVPR (2019)

\bibitem{singh2019finegan}
Singh, K.K., Ojha, U., Lee, Y.J.: Finegan: Unsupervised hierarchical
  disentanglement for fine-grained object generation and discovery. In: CVPR
  (2019)

\bibitem{tang2019attention}
Tang, H., Xu, D., Sebe, N., Yan, Y.: Attention-guided generative adversarial
  networks for unsupervised image-to-image translation. In: International Joint
  Conference on Neural Networks (IJCNN) (2019)

\bibitem{tang2020multi}
Tang, H., Xu, D., Yan, Y., Corso, J.J., Torr, P.H., Sebe, N.: Multi-channel
  attention selection gans for guided image-to-image translation. In: CVPR
  (2019)

\bibitem{uricar2019let}
Uricar, M., Sistu, G., Rashed, H., Vobecky, A., Krizek, P., Burger, F.,
  Yogamani, S.: Let's get dirty: Gan based data augmentation for soiling and
  adverse weather classification in autonomous driving. arXiv preprint
  arXiv:1912.02249  (2019)

\bibitem{xiao2017dna}
Xiao, T., Hong, J., Ma, J.: Dna-gan: Learning disentangled representations from
  multi-attribute images. ICLR Workshops  (2018)

\bibitem{xiao2018elegant}
Xiao, T., Hong, J., Ma, J.: Elegant: Exchanging latent encodings with gan for
  transferring multiple face attributes. In: ECCV (2018)

\bibitem{xie2018tempogan}
Xie, Y., Franz, E., Chu, M., Thuerey, N.: tempogan: A temporally coherent,
  volumetric gan for super-resolution fluid flow. SIGGRAPH  (2018)

\bibitem{yang2018towards}
Yang, X., Xu, Z., Luo, J.: Towards perceptual image dehazing by physics-based
  disentanglement and adversarial training. In: AAAI (2018)

\bibitem{yang2018crossing}
Yang, X., Xie, D., Wang, X.: Crossing-domain generative adversarial networks
  for unsupervised multi-domain image-to-image translation. In: MM (2018)

\bibitem{yi2017dualgan}
Yi, Z., Zhang, H., Tan, P., Gong, M.: Dualgan: Unsupervised dual learning for
  image-to-image translation. In: ICCV (2017)

\bibitem{yogamani2019woodscape}
Yogamani, S., Hughes, C., Horgan, J., Sistu, G., Varley, P., O'Dea, D.,
  Uric{\'a}r, M., Milz, S., Simon, M., Amende, K., et~al.: Woodscape: A
  multi-task, multi-camera fisheye dataset for autonomous driving. In: ICCV
  (2019)

\bibitem{you2015adherent}
You, S., Tan, R.T., Kawakami, R., Mukaigawa, Y., Ikeuchi, K.: Adherent raindrop
  modeling, detectionand removal in video. T-PAMI  (2015)

\bibitem{zhang2019multi}
Zhang, J., Huang, Y., Li, Y., Zhao, W., Zhang, L.: Multi-attribute transfer via
  disentangled representation. In: AAAI (2019)

\bibitem{zhang2018unreasonable}
Zhang, R., Isola, P., Efros, A.A., Shechtman, E., Wang, O.: The unreasonable
  effectiveness of deep features as a perceptual metric. In: CVPR (2018)

\bibitem{zhao2017pyramid}
Zhao, H., Shi, J., Qi, X., Wang, X., Jia, J.: Pyramid scene parsing network.
  In: CVPR (2017)

\bibitem{zhu2017unpaired}
Zhu, J.Y., Park, T., Isola, P., Efros, A.A.: Unpaired image-to-image
  translation using cycle-consistent adversarial networks. In: CVPR (2017)

\bibitem{zhu2017toward}
Zhu, J.Y., Zhang, R., Pathak, D., Darrell, T., Efros, A.A., Wang, O.,
  Shechtman, E.: Toward multimodal image-to-image translation. In: NeurIPS
  (2017)

\end{thebibliography}

\end{document}